\documentclass[11pt]{article}

\usepackage[preprint]{acl}

\usepackage{times}
\usepackage{latexsym}

\usepackage[T1]{fontenc}

\usepackage[utf8]{inputenc}

\usepackage{microtype}

\usepackage{inconsolata}

\usepackage{amsmath}
\usepackage{graphicx}
\usepackage[most]{tcolorbox}
\usepackage{multirow}
\usepackage{tcolorbox}
\usepackage{algorithm}

\definecolor{tabblue}{HTML}{1F77B4}
\definecolor{taborange}{HTML}{FF7F0E}
\definecolor{tabgreen}{HTML}{2CA02C}
\definecolor{tabred}{HTML}{D62728}
\definecolor{tabpurple}{HTML}{9467BD}
\definecolor{tabbrown}{HTML}{8C564B}
\definecolor{tabpink}{HTML}{E377C2}
\definecolor{tabgray}{HTML}{7F7F7F}
\definecolor{tabolive}{HTML}{BCBD22}
\definecolor{tabcyan}{HTML}{17BECF}

\usepackage{rotating} 
\usepackage[most]{tcolorbox}
\newtcolorbox{takeaway}{
  colback=blue!5,
  colframe=blue!5,
  width=\columnwidth,
  boxrule=0pt,
  left=5pt, right=5pt, top=3pt, bottom=3pt, 
  boxsep=0pt, 
}
      
\PassOptionsToPackage{table}{xcolor}
\usepackage{xcolor}      

\usepackage{wrapfig}
\usepackage{floatflt}
\usepackage{color, colortbl}

\usepackage[inline,shortlabels]{enumitem}

\usepackage{lineno}

\definecolor{darkblue}{rgb}{0, 0, 0.5}
\hypersetup{colorlinks=true, citecolor=darkblue, linkcolor=darkblue, urlcolor=darkblue}
\definecolor{Highlight}{rgb}{0.92,0.94,1}

\usepackage{tabularx}
\PassOptionsToPackage{table}{xcolor}
\usepackage{booktabs}
\usepackage{wrapfig}
\usepackage{graphicx}
\usepackage{amsfonts} 
\usepackage{float}
\usepackage{color, colortbl}

\title{Bridging Reasoning Trajectories in On-Policy Distillation via Near-Future Guidance}

\author{
Yuxuan Jiang$^{1}$ \quad Francis Ferraro$^{1}$ \\
$^{1}$University of Maryland, Baltimore County \\
\texttt{yuxuanj1@umbc.edu}
}

\begin{document}
\maketitle

\begin{abstract}
On-Policy Distillation (OPD) improves large language model reasoning by training a student on trajectories sampled from its own policy under teacher supervision. Although OPD operates on trajectories, its learning signal remains token-level: it identifies deviations through high-loss tokens and repairs them through local reverse-KL correction. We show that this ``trajectory-sampled but token-learned'' mechanism cannot reliably bridge student trajectories toward teacher trajectories. About 30\% of high-loss tokens fall into the low-divergence regime, indicating that many are surface-form mismatches rather than real reasoning forks. Moreover, even truly divergent tokens are difficult to repair with isolated token-level supervision, since reasoning failures often unfold as short-horizon distributional drift. We propose \emph{Trajectory-aware OPD} (TOPD), which uses near-future trajectory information to identify real divergent states and distribute guidance across multiple future tokens. Experiments show that suppressing non-divergent high-loss tokens improves standard OPD from 47.8\% to 48.2\% average accuracy, while TOPD further improves performance to 52.2\%, with gains on AIME24 from 60.0\% to 63.3\% and AIME25 from 46.7\% to 53.3\%. 
\end{abstract}

\section{Introduction}
On-Policy Distillation (OPD) has established itself as a cornerstone of the modern post-training pipeline for Large Language Models (LLMs)~\cite{agarwal2024policy,tan2024large}. This effectiveness has been validated by industrial works such as DeepSeek-V4~\cite{deepseekv4}, MiMo~\cite{xiao2026mimo}, and Qwen-3~\cite{qwen3technicalreport}, where OPD serves as a vital component alongside Supervised Fine-tuning (SFT) or Reinforcement Learning with Verifiable Rewards (RLVR) \citep{dipta2026ganitllm} to further squeeze out reasoning performance.

\begin{figure*}

  \centering
  \includegraphics[width=\linewidth]{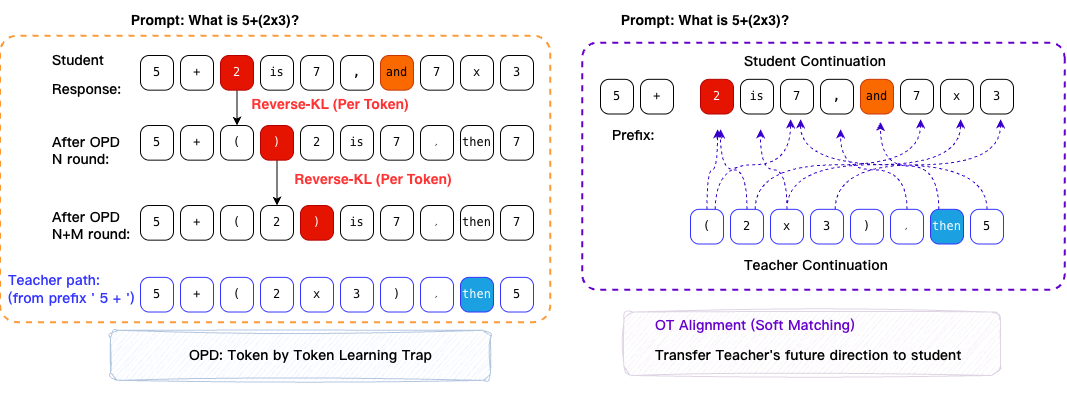}
\caption{
Overview of standard OPD and TOPD. Left: a high-loss token such as ``and'' may be a noisy candidate, and reverse-KL OPD can fall into the token-by-token learning trap where local token correction fails to repair the future trajectory. Right: TOPD uses short-window OT alignment to inject trajectory-level guidance and move the student continuation toward the teacher path.
}

  \label{fig:intro}
\end{figure*}

OPD offers a natural way to improve reasoning models by supervising the student on trajectories sampled from its own policy. Since these trajectories reflect the states that the student actually visits, high-loss tokens provide useful signals about where the student may deviate from the teacher. Standard reverse-KL correction then encourages the student to move away from these low-probability actions and toward the teacher-preferred behavior~\cite{lu2025onpolicydistillation}. This token-level mechanism makes OPD an effective and practical approach for refining reasoning trajectories.

However, we find that OPD cannot reliably bridge student reasoning trajectories toward teacher trajectories. As illustrated in Figure~\ref{fig:intro}, OPD suffers from two closely related failure modes in trajectory-level reasoning correction: high-loss tokens may correspond to false alarms, and more critically, the model can become trapped in a token-by-token learning process where local corrections fail to restore the overall reasoning path.

First, we find that high-loss tokens do not always correspond to real reasoning divergence. Although high-loss tokens reflect strong disagreement between teacher and student at local predictions, this does not necessarily mean that the two models will follow highly divergent reasoning trajectories from that point onward. For example, the token ``and'' in Figure~\ref{fig:intro} incurs high loss mainly because the teacher prefers connective expressions such as ``then,'' while the subsequent reasoning process remains nearly identical. Our short-window probing results show that token-level loss is only weakly aligned with near-future trajectory divergence, and a substantial fraction of high-loss tokens are actually low-divergence false alarms. Moreover, ignoring these false alarms improves OPD performance, suggesting that they not only fail to provide useful supervision, but can actively interfere with reasoning correction.

More importantly, even after identifying real divergent points, single-token reverse-KL correction still struggles to repair an entire reasoning trajectory. Multi-step reasoning failures rarely appear as isolated token mistakes; instead, they gradually evolve into distributional drift over a short future window. Figure~\ref{fig:intro} illustrates this token-by-token learning trap: in the example of \(5+2\times3\), OPD gradually changes local tokens from the incorrect ``2'' toward ``('' and further adjusts nearby symbols, yet the model still continues to generate ``is 7,'' indicating that the underlying reasoning trajectory has not truly returned to the teacher-guided path \(5+(2\times3)\). In other words, local token correction does not efficiently reshape the student's near-future transition distribution under the same prefix.

Motivated by these observations, we propose \emph{Trajectory-aware OPD (TOPD)}. TOPD leverages near-future trajectory information to build a bridge between local token correction and reasoning trajectory evolution, helping OPD overcome the mismatch between token-level learning and trajectory-level reasoning correction. Concretely, TOPD first compares teacher and student short-window continuations generated from the same prefix to identify genuinely trajectory-divergent states and filter out high-loss false alarms. It then uses OT-based trajectory alignment to transfer future trajectory direction into the learning objective, allowing the student to learn not only how to correct the current token, but also how its subsequent reasoning trajectory should move toward the teacher path. In this way, OPD supervision is extended from isolated token correction to short-window trajectory correction, enabling more direct optimization of reasoning trajectories.
Our contributions are summarized as follows:
\begin{itemize}
    \item This work identifies a fundamental mismatch between token-level supervision and trajectory-level reasoning correction in OPD. Although OPD is trained with token-level reverse-KL supervision, multi-step reasoning requires trajectory-level error detection and path repair, which token-level supervision does not reliably provide.

    \item Our empirical analysis reveal two failure mechanisms caused by this mismatch. Our probing results show that about 30\% of high-loss tokens are actually low-divergence false alarms, indicating that high loss does not always correspond to real trajectory divergence. We further identify the token-by-token learning trap, where single-token correction fails to sufficiently repair future reasoning trajectories.

    \item We introduce a trajectory-level guidance principle that injects future trajectory information into the losses of multiple tokens. Instead of restricting supervision to local token correction, OPD should distribute the teacher's near-future trajectory information across a short-window training objective. Based on this principle, short-window OT alignment is used to inject teacher-student path discrepancy into the loss, encouraging the student's future reasoning path to move toward the teacher trajectory.
\end{itemize}

\section{When Token-Level OPD Fails to Redirect Reasoning Trajectories}

In principle, On-Policy Distillation aims to align the student's reasoning \textit{trajectory} with the teacher's. While formulated through token-level loss, the success of OPD implicitly hinges on two fundamental capabilities: the ability to \textit{identify} where the student has diverged, and the ability to \textit{correct} the reasoning path thereafter. \textbf{However, we identify two critical failure modes in standard OPD that suggest token-level supervision is structurally misaligned with these requirements.}

\subsection{Informative but Blunt: High-Loss Tokens Are Not Always Trajectory-Critical}
\label{motiv1}

\begin{figure}

  \centering
  \includegraphics[width=\linewidth]{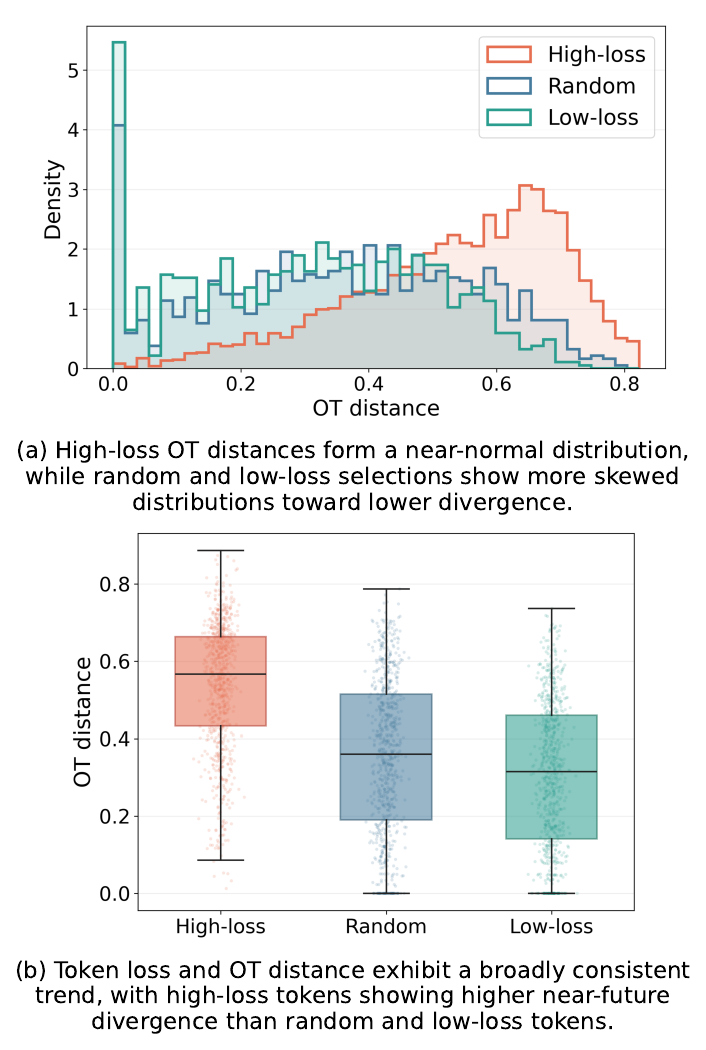}
\caption{
Near-future divergence analysis. High-loss tokens show larger OT distances on average, but substantial overlap with random and low-loss tokens indicates that token-level loss is an informative yet noisy indicator of trajectory divergence.
}

  \label{fig:divergence}
\end{figure}

A common heuristic in existing On-Policy Distillation is that high-loss tokens mark trajectory-critical divergence points requiring strong correction. However, this assumption is often fragile. We observe that high token loss frequently captures benign surface-level mismatches—such as stylistic preferences or equivalent symbolic forms—rather than genuine logical deviations. Consequently, a high-loss token may signal a local teacher-student mismatch without implying a divergent future reasoning path.

To examine this phenomenon, we conduct a short-window probing analysis using \textbf{Qwen3-30B-A3B-Instruct-2507} as the teacher and \textbf{Qwen3-4B-Instruct-2507} as the student, student trajectories are generated from prompts sampled from \textbf{OpenThoughts3}. For each trajectory, we compare the near-future continuations induced by high-loss, random, and low-loss token positions, and use short-window (length $K=50$ tokens) OT distance as a trajectory-level divergence measure.Figure~\ref{fig:divergence} shows that high-loss tokens indeed exhibit substantially larger OT distances on average, suggesting that token-level loss broadly reflects future trajectory divergence. Specifically, the median OT distance of high-loss tokens is 0.566, compared with 0.361 for random tokens and 0.315 for low-loss tokens. This indicates that the core intuition behind OPD---using high-loss tokens as correction targets---is statistically meaningful.

However, token-level loss remains a weak and noisy predictor of actual trajectory divergence. The correlation between token loss and near-future OT distance is limited (Pearson \(r=0.126\), Spearman \(\rho=0.143\)), and a substantial overlap exists between high-loss tokens and low-divergence regions (more details in Appendix~\ref{correlation}). In particular, approximately 29.23\% of high-loss tokens fall within the typical low-divergence range defined by the upper quartile of low-loss OT distances. These tokens correspond to local stylistic or surface-form mismatches that do not substantially alter the subsequent reasoning trajectory.

\begin{takeaway}
\textcolor{tabblue}{\textbf{Take-away}}~
While token-level loss serves as a useful first-order indicator of policy discrepancy, it remains a remarkably \textit{blunt proxy} that lacks structural precision. A substantial fraction of high-loss tokens are statistically indistinguishable from benign, low-divergence steps, blurring the boundary between genuine reasoning forks and harmless local variation.
\end{takeaway}

\subsection{Local Token Correction Does Not Guarantee Trajectory Correction}

More importantly, even when a genuine divergence point is identified, imposing a strong reverse-KL penalty on a single token is often insufficient to redirect the reasoning trajectory. Such localized supervision primarily encourages the model to match the teacher's distribution at the current step, without ensuring that subsequent generations align with the teacher’s intended reasoning path. In short, while existing OPD methods can identify \emph{where} a student diverges, token-level supervision remains agnostic to \emph{how} the student should navigate the reasoning trajectory afterward.

To illustrate this issue, we design a local trajectory correction setting. For each trajectory, we first identify one divergent point using the procedure described above. We then step back by one position and fix the original prompt together with the student generation before this position as the prefix. From the same prefix, we regenerate a fixed-length (50 Tokens) continuation and compute the reverse-KL OPD loss only on this continuation for model update. After the update, the model regenerates a continuation from the same prefix, and we compare its trajectory divergence from the teacher continuation. To make the local effect of reverse-KL optimization easier to observe, we perform multiple consecutive updates on the same prefix.

\begin{figure}

  \centering
  \includegraphics[width=\linewidth]{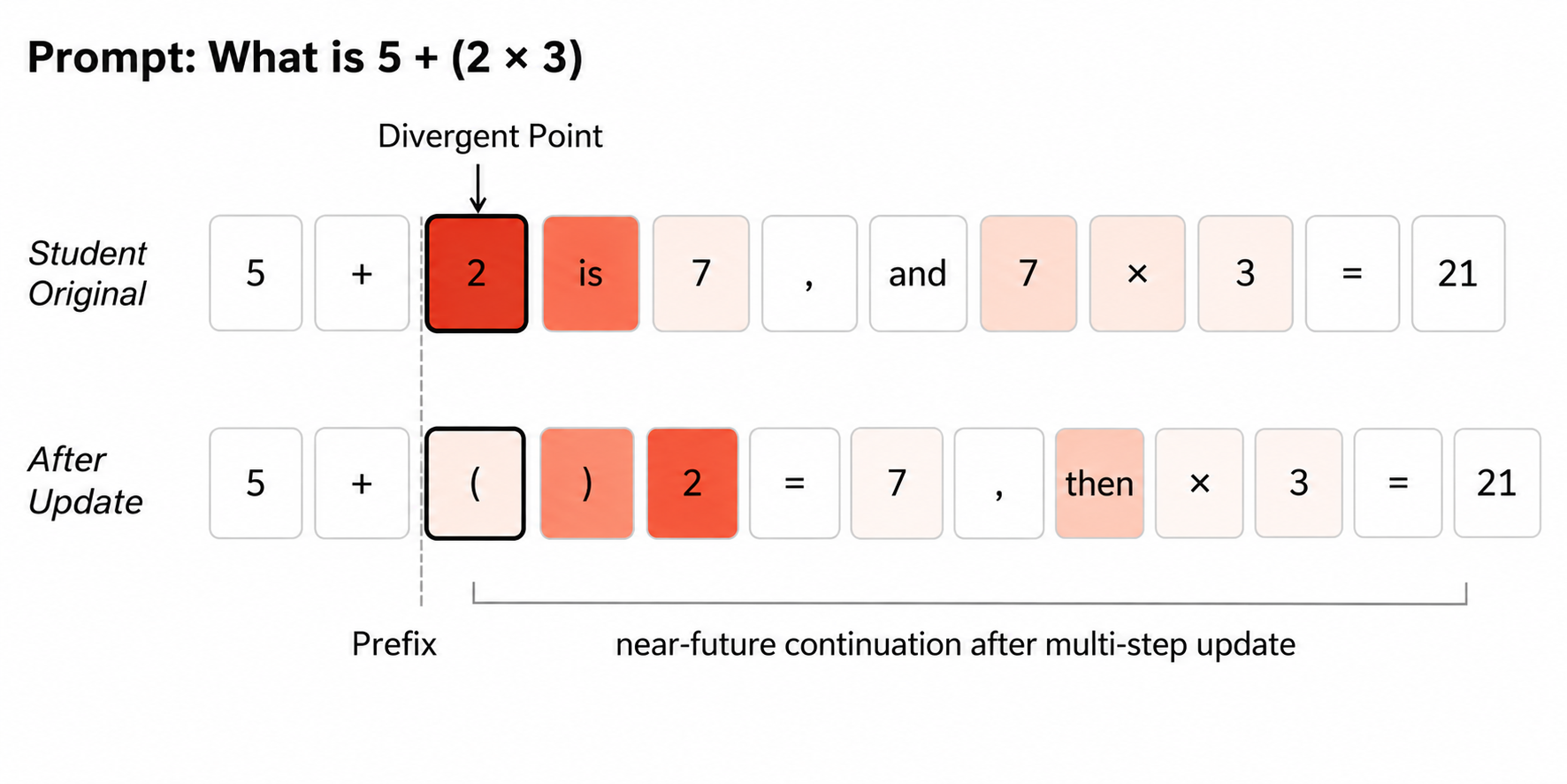}
\caption{
Case study of the token-by-token learning trap. The teacher expects the correction ``('' to guide the student toward \(5+(2\times3)=11\), but the student instead enters another erroneous branch, showing that local token correction does not guarantee trajectory redirection.
}

  \label{fig:case}
\end{figure}

\paragraph{Case study.}
Figure~\ref{fig:case} illustrates a local correction failure. Given the prompt ``What is \(5 + (2 \times 3)\)?'', the current prefix is \(5+\). At this position, the teacher-expected continuation is \((2\times3)=6\), so the next token should be the left parenthesis ``(''; this also serves as the beginning of the teacher's own future trajectory. In contrast, the student's original generation starts with the token ``2'' and continues along its erroneous path by computing \(5+2=7\), eventually reaching \(7\times3=21\).

After the local OPD update, the student is indeed corrected at the divergent position and generates the teacher-expected left parenthesis ``(''. However, this local correction does not change the subsequent transition distribution. At the next position, the student fails to continue with the teacher-expected \(2\times3\) structure; instead, it generates ``)'' and then returns to its original erroneous path, producing \(2=7\). In other words, OPD successfully changes the current token, but it does not make the model enter the teacher-consistent near-future trajectory.

This case reveals what we call the \textit{token-by-token learning trap}. OPD successfully corrects the current divergent token, but the subsequent continuation still fails to enter the teacher-consistent trajectory. Instead, the model drifts into a new erroneous branch, where future tokens such as ``)'' and the following continuation would require additional rounds of local correction. As a result, reverse-KL supervision repairs the trajectory only by editing one position at a time, rather than directly inducing a coherent trajectory-level shift toward the teacher path. This suggests that local reverse-KL supervision changes individual token preferences, but does not sufficiently reshape the model's near-future transition dynamics under the corrected prefix.

\begin{takeaway}
\textcolor{tabblue}{\textbf{Take-away}}~
\textbf{Toke by Token Learning Trap} : treating reasoning correction as a series of independent token-level targets is insufficient for trajectory-level alignment. As long as supervision remains point-wise, models fail to escape erroneous reasoning branches, as they lack the global perspective to rectify the subsequent \textit{near-future reasoning drift}. Meaningful trajectory repair necessitates shifting from local token editing to global, transition-aware alignment.
\end{takeaway}

\section{Methodology}

We propose \textbf{Trajectory Aware OPD (TOPD)}. TOPD first identifies \emph{real divergent points} that induce near-future reasoning drift, and then injects short-window teacher-student trajectory discrepancy into the training objective. This enables the student to learn not only which token to correct, but also how to move toward the teacher's near-future reasoning trajectory.

\paragraph{Divergence Detection.}
Given a teacher model and a student model, we first compute the token-level OPD loss along the student-generated trajectory. For each high-loss candidate position \(t\), we step back to \(t-1\) and keep the student prefix \(x^S_{<t}\). Starting from this same prefix, both the teacher and student generate a continuation of length \(K\). Since the prefix ends at \(t-1\), the future window starts from position \(t\) and includes the candidate token itself. We denote the two short-window trajectories as
\[
T_{t:t+K}
=
\{x^T_{t+i}\}_{i=0}^{K-1},
\quad
S_{t:t+K}
=
\{x^S_{t+i}\}_{i=0}^{K-1}.
\]

We map both continuations into the embedding space and measure their trajectory-level discrepancy using optimal transport. Let
\[
C_{ij}=c(x^T_{t+i},x^S_{t+j})
\]
be the ground-cost matrix between teacher and student future tokens. The short-window OT distance is defined as
\[
D_{\mathrm{OT}}(T_{t:t+K}, S_{t:t+K})
=
\min_{\gamma \in \Pi(a,b)}
\sum_{i=0}^{K-1}
\sum_{j=0}^{K-1}
\gamma_{ij} C_{ij},
\]
where \(\gamma\) is the transport plan, and \(\Pi(a,b)\) denotes the set of admissible transport matrices satisfying the marginal constraints \(a\) and \(b\). We treat positions with both high token-level loss and high short-window OT divergence as real divergent points.

\paragraph{Trajectory-level Teacher Signal Injection.}
After detecting a real divergent point \(t\), TOPD extends supervision from the current token to the short-window trajectory generated from the same prefix \(x^S_{<t}\). Let
\[
X_T
=
\{x^T_{t+i}\}_{i=0}^{K-1},
\quad
X_S
=
\{x^S_{t+i}\}_{i=0}^{K-1}
\]
denote the teacher and student future trajectories. Since both are generated from the same prefix, their discrepancy reflects how the two models choose different near-future reasoning paths under the same state.

Using the OT transport plan \(\gamma\), we construct a trajectory-aware soft target for each student future position. Specifically, \(\gamma_{ij}\) represents the soft alignment between the \(i\)-th student position and the \(j\)-th teacher position. The target for the \(i\)-th student future position is
\[
\tilde{y}_{t+i}
=
\sum_{j=0}^{K-1}
\gamma_{ij}
\cdot
\mathrm{onehot}(x^T_{t+j}).
\]

We then train the student prediction distribution \(p_S(\cdot \mid c_{t+i})\) to match this OT-aligned soft target:
\[
\mathcal{L}_{\mathrm{traj}}
=
\sum_{i=0}^{K-1}
\mathrm{KL}
\left(
\tilde{y}_{t+i}
\parallel
p_S(\cdot \mid c_{t+i})
\right),
\]
where \(c_{t+i}\) is the context for the \(i\)-th student future state. The final TOPD objective is
\[
\mathcal{L}_{\mathrm{TOPD}}
=
\mathcal{L}_{\mathrm{OPD}}
+
\lambda
\mathcal{L}_{\mathrm{traj}},
\]
where \(\lambda\) controls the strength of trajectory-level supervision.

In this way, TOPD preserves the original token-level correction signal while adding near-future trajectory guidance. Rather than only aligning the current token, the student is encouraged to follow a teacher-consistent short-window reasoning path, which helps alleviate the token-by-token learning trap.

\section{Empirical Experiments}

\label{setup}
\subsection{Training Setup}

We use KDFlow~\cite{zhang2026kdflow} to distill \textbf{Qwen3-30B-A3B-Instruct-2507} into \textbf{Qwen3-4B-Instruct-2507}, with thinking mode disabled for both models.

Training consists of two stages.
In Stage 1, the student is initialized via off-policy distillation on 20k teacher-generated solutions from \textbf{OpenThoughts3}~\cite{guha2025openthoughts} using forward KL distillation combined with cross-entropy (\(\texttt{kd\_ratio}=0.5\)).

In Stage 2, the student performs on-policy sampling on a separate set of 50k prompts. For each prompt, we sample 4 student rollouts and optimize the student on its own trajectories using reverse KL distillation (\(\texttt{kd\_ratio}=1.0\)).

Unless otherwise specified, all remaining hyperparameters follow the default KDFlow configuration. Additional implementation details are provided in Appendix~\ref{app:training_details}.

\paragraph{Computation Cost}
In practice, TOPD introduces only a modest overhead: OT is computed only on selected high-loss short windows, and the overall training time is \(1.41\times\) that of standard OPD in our experiments.

\subsection{Evaluation}
We employ the \textsc{lm-evaluation-harness} framework \cite{eval-harness} for standardized assessment across all benchmarks in a zero-shot setting. To ensure statistical robustness, we report the \textbf{Pass@1} accuracy averaged over five independent runs, accounting for variance in decoding.

Our evaluation suite focuses on challenging competitive mathematics, comprising \textbf{AIME 24}, \textbf{AIME 25}, and \textbf{HMMT 25-Feb} from MathArena~\cite{dekoninck2026matharena}. Each benchmark contains 30 problems; we report the aggregated average score across these 90 tasks as the primary performance metric in our study.

\subsection{Main Results}

\begin{table*}[t]
\centering
\small
\begin{tabular}{lcccc}
\toprule
Method & AIME24 & AIME25 & HMMT25-Feb & Avg. \\
\midrule
Qwen3-4B (Warm Start) & 46.7 & 40.0 & 30.0 & 38.9 \\
+ OPD & 60.0 & 46.7 & 36.7 & 47.8 \\
\rowcolor{blue!6}
+ TOPD (Ours) & \textbf{63.3} & \textbf{53.3} & \textbf{40.0} & \textbf{52.2} \\
\bottomrule
\end{tabular}
\caption{
Main results on competitive mathematics benchmarks. TOPD consistently improves standard OPD by incorporating trajectory-aware future guidance beyond point-wise token-level correction.
}
\label{tab:main_result}
\end{table*}

Table~\ref{tab:main_result} presents the main results on three competitive mathematics benchmarks. The offline-distilled Qwen3-4B model achieves an average accuracy of 38.9. Standard OPD substantially improves the performance to 47.8, demonstrating the effectiveness of on-policy reasoning distillation. TOPD further improves the average accuracy to 52.2 and consistently outperforms standard OPD across all benchmarks.

In particular, TOPD improves AIME24 from 60.0 to 63.3 and AIME25 from 46.7 to 53.3, showing that trajectory-aware future supervision is more effective than point-wise token-level correction alone. These results support our hypothesis that reasoning failures are fundamentally trajectory-level phenomena, and that injecting near-future trajectory guidance can more effectively redirect the student toward teacher-consistent reasoning paths.

\section{Ablations and Analysis}

\subsection{Trajectory-Aware Divergence Detection}

As shown in the previous section~\ref{motiv1}, not all high-loss tokens correspond to real trajectory-level divergence. Here, we further examine whether these high-loss false alarms have a measurable impact on OPD training. Specifically, we test whether suppressing low-OT high-loss tokens (\emph{false alarms}) improves OPD, and whether suppressing high-OT high-loss tokens (\emph{real divergent points}) removes useful correction signals. 

\begin{figure}[t]
  \centering
  \includegraphics[width=\linewidth]{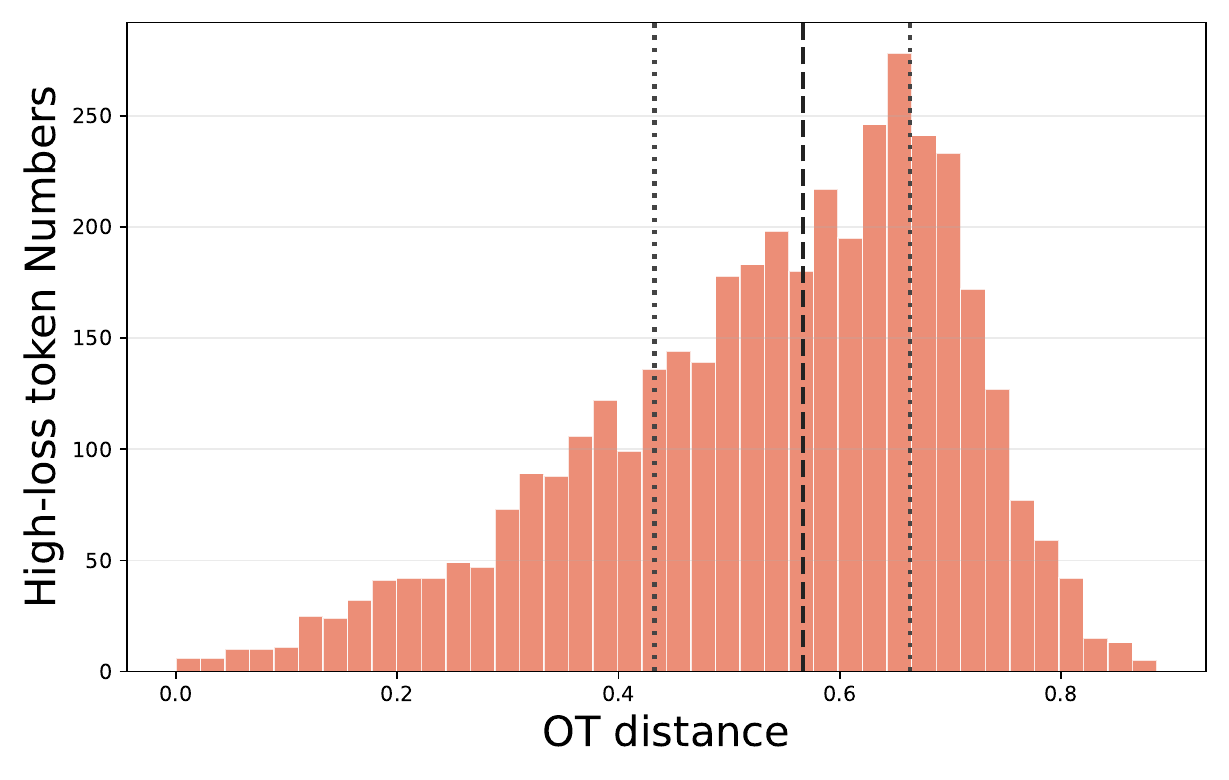}
  \caption{
  OT distribution of high-loss tokens. Although high-loss tokens exhibit larger trajectory divergence on average, their OT distances span a broad range, indicating that many high-loss positions remain near the low-divergence regime rather than corresponding to genuine reasoning forks.
  }
  \label{fig:high}
\end{figure}

\paragraph{Downweighting strategy.}
Figure~\ref{fig:high} shows that high-loss tokens span a wide range of OT distances rather than concentrating only in the highly divergent region. This motivates a targeted downweighting intervention: if low-OT high-loss tokens are noisy supervision signals, suppressing them should not harm OPD and may improve training; in contrast, suppressing high-OT high-loss tokens should remove genuinely useful correction signals. For each training sample, we compute the median token loss within the sample as a local reference level. For selected tokens, we cap their contribution at this sample-level median loss instead of using their original high loss.

\paragraph{Variants.}
We compare three settings: (1) \textbf{Baseline OPD}, which applies standard OPD without additional downweighting; (2) \textbf{Low-OT High-Loss Downweighting}, which downweights the bottom 30\% OT tokens among high-loss tokens; and (3) \textbf{Matched High-OT High-Loss Downweighting}, which downweights the same number of tokens randomly sampled from the high-OT region of high-loss tokens.

\begin{table}[t]
\centering
\small
\begin{tabular}{lc}
\toprule
Method & Avg.Acc \\
\midrule
\textbf{Qwen3-4B (Warm Start)} & 38.9 \\
+ Baseline OPD & 47.8 \\
\textbf{+ Low-OT High-Loss DW} & \textbf{48.2} \\
+ Matched High-OT DW & 42.1 \\
\rowcolor{blue!6}
+ TOPD & \textbf{52.2} \\
\bottomrule
\end{tabular}
\caption{
Mechanistic analysis of divergence-aware supervision. Suppressing low-OT high-loss tokens slightly improves OPD, whereas suppressing matched high-OT tokens substantially harms performance. TOPD achieves the best result by further injecting trajectory-aware future guidance.
}
\label{tab:short_opd}
\end{table}

\paragraph{Results and findings.}
Table~\ref{tab:short_opd} shows that standard OPD improves the warm-start Qwen3-4B model from 38.9\% to 47.8\%. Downweighting low-OT high-loss tokens further improves accuracy to 48.2\%, suggesting that many low-divergence high-loss positions act as noisy supervision signals. In contrast, downweighting matched high-OT tokens drops performance to 42.1\%, indicating that high-OT high-loss positions indeed contain important trajectory-critical correction signals. This contrast confirms that trajectory-aware divergence detection helps distinguish benign local mismatch from real reasoning divergence. Finally, TOPD achieves the best accuracy of 52.2\%, showing that identifying real divergent points alone is not sufficient; injecting near-future trajectory guidance provides an additional and substantially larger gain.

\begin{takeaway}
\textcolor{tabblue}{\textbf{Take-away}}~
By suppressing the bottom 30\% low-OT high-loss tokens, i.e., high-loss false alarms, OPD achieves a further +0.4\% improvement over standard OPD. This suggests that these false alarms are not merely uninformative, but can actively hinder OPD training. Moreover, our results indicate that effective OPD requires not only accurate identification of real divergent points, but also trajectory-aware future guidance for repairing reasoning paths.
\end{takeaway}

\begin{figure}[t]
  \centering
  \includegraphics[width=\linewidth]{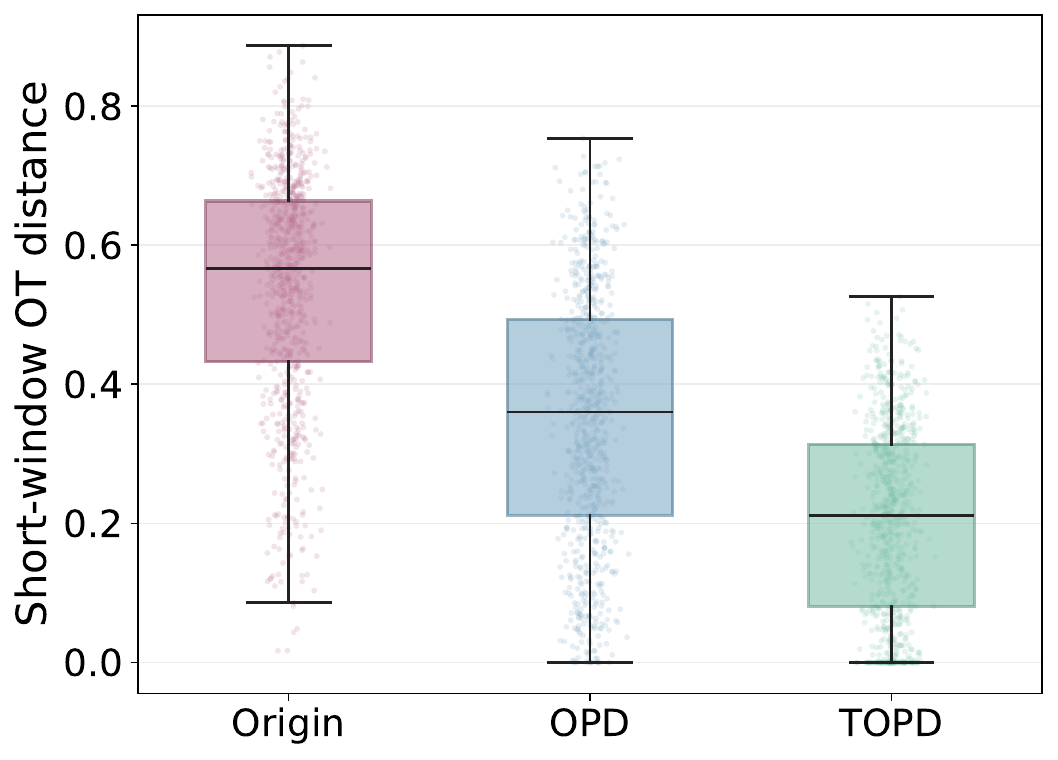}
  \caption{
  Local trajectory correction analysis. We compare the short-window OT distance before local updates, after standard OPD updates, and after TOPD updates on the same divergent prefixes. Each point represents one local correction sample. Standard OPD reduces the trajectory divergence only partially, while TOPD shifts the OT distribution substantially further downward, indicating stronger near-future trajectory repair toward the teacher continuation.
  }
  \label{fig:local_traj}
\end{figure}
\subsection{Does TOPD Improve Local Trajectory Correction?}

The main results show that TOPD improves OPD under the reverse-KL training setting. We further ask whether this performance gain is accompanied by the intended trajectory-level improvement: after a real divergent point, does TOPD more effectively move the student's near-future continuation toward the teacher trajectory? To answer this question, we design a local trajectory correction experiment that directly measures short-window OT divergence before and after local updates.

\paragraph{Local Trajectory Correction.} For each of 1,000 training samples, we select one real divergence point that exhibits both high token-level loss and high short-window OT divergence. We then concatenate the original problem with the student response before this divergence point to form a fixed prefix. Starting from this same prefix, we use the teacher continuation as the local optimization target and retain only the next 50 tokens after the divergence point as the correction window. Tokens outside this window are masked out and do not contribute to the gradient update.

We compare two local correction strategies. The first is standard OPD, which optimizes the token-level distillation loss within the short window. The second is TOPD, which applies the same short-window correction setting but additionally injects trajectory-aware future guidance through the OT-based loss term. To make the local learning effect observable, we perform \(M=5\) consecutive gradient updates on the same prefix-window pair.

After the local updates, we regenerate a continuation from the same fixed prefix using the updated model. We then compute the short-window OT distance between the regenerated student continuation and the teacher continuation. By comparing the OT distance before and after local updates, we measure whether each method successfully moves the student's near-future reasoning trajectory toward the teacher trajectory.

\paragraph{Results and findings.}
Figure~\ref{fig:local_traj} shows that divergent prefixes have a high initial short-window OT distance, with a mean of 0.538. Standard OPD reduces the mean OT distance to 0.350, suggesting that token-level reverse-KL supervision can partially repair local trajectory drift. However, TOPD further reduces the mean OT distance to 0.204 and produces a more concentrated low-OT distribution. This experiment suggests that the performance gain of TOPD comes from enabling the student to more effectively absorb and follow the teacher's near-future trajectory information during local correction.

\section{Related Works}
\subsection{On-Policy Distillation for LLM Post-Training}

On-policy distillation (OPD) evolves beyond SFT and RLVR by combining trajectory exploration with dense, token-level guidance~\cite{lu2025onpolicydistillation}. Recent large-scale reasoning systems such as DeepSeek-V4~\cite{deepseekv4}, MiMo~\cite{xiao2026mimo}, and Qwen-3~\cite{qwen3technicalreport} demonstrate the effectiveness of OPD as a post-training paradigm, while self-distillation studies suggest that OPD can amplify latent reasoning capabilities rather than merely imitate teacher outputs~\cite{zhao2026self,shenfeld2026self,hubotter2026reinforcement}. 

Recent analyses further reveal that successful OPD depends on several factors, including complementary teacher reasoning patterns, sufficient reasoning-style overlap between teacher and student, and adaptive teacher selection strategies~\cite{li2026rethinking,fu2026revisiting,jiang2026cornerstones}. At the same time, growing attention has been paid to reasoning efficiency~\cite{shen2026efficiency} and compression during post-training. Existing studies explore structured pruning, reasoning compression, and efficiency-aware optimization to reduce unnecessary reasoning steps while preserving performance~\cite{jiang2025drp,11460600,li2024sglp,10800533}. Related work additionally investigates performance-efficiency trade-offs across different model scales and task settings~\cite{cao2026taskspecificefficiencyanalysissmall,zhang2026performance}. 

Beyond pure optimization, recent work also highlights broader limitations of current reasoning systems, including memorization-constrained reasoning beyond mathematical benchmarks~\cite{jiang2026beyond} and the importance of intermediate reasoning structure during tool-integrated optimization~\cite{Li2025Efficient,xu2025learning,al2026dagger} and the role of explicit reasoning decomposition \citep{roy-dipta-ferraro-2025-may,roy-dipta-ferraro-2025-q2e}. Despite these advances, most existing OPD methods still rely on reverse-KL objectives defined at the token level. Such formulations provide limited modeling capacity for cross-step dependencies and long-horizon trajectory consistency. Our work instead explicitly incorporates trajectory-aware path information into distillation, enabling more effective reasoning alignment.

\subsection{Optimal Transport and Structured Alignment}

Existing distillation methods typically rely on forward-KL or reverse-KL for distribution alignment. While forward-KL tends to cover the overall teacher distribution~\cite{zhu2026hybrid,li2025ammkd}, OPD commonly adopts reverse-KL to correct student exploration by penalizing tokens that substantially deviate from teacher preferences~\cite{lu2025onpolicydistillation,hou2026uniopdunifyingonpolicydistillation}. However, such token-level objectives usually treat each prediction step independently, making it difficult to capture cross-step dependencies in reasoning trajectories~\cite{lv2024wasserstein}. 

To address these limitations, prior studies introduce optimal transport (OT) and structure-aware matching objectives for modeling discrepancies between teacher and student distributions~\cite{bhardwaj2022knot,cui2025multi,luo2025cellink}. Similar alignment ideas have also been explored in broader representation learning and retrieval settings, including frequency- and spectral-aligned distillation strategies~\cite{li2025frequency,11460474}, preference-aware optimization~\cite{li2025preference}, and multimodal or multi-teacher alignment frameworks~\cite{li2025mmt,zhang2025find,li2025srkd}. 

Recent multimodal retrieval studies further emphasize the importance of structured semantic grounding and compositional alignment~\cite{miao2026seeing}. Existing work investigates explicit semantic parsing and entity-aware representation learning for compositional retrieval~\cite{li2025ddtime,FineCIR,ENCODER}, while robust alignment under complex modification signals motivates progressive learning and noise-unlearning frameworks~\cite{HABIT,ConeSep,xu2026stable}. In parallel, anchor-based calibration mechanisms have been explored in both image and video retrieval settings~\cite{TEMA,ReTrack}, and arbiter-calibrated retrieval strategies provide another perspective on robust semantic alignment~\cite{Air-Know}. Related representation learning approaches additionally study comprehensive attribute exploration for zero-shot retrieval and hashing~\cite{li2024comae,wang2026tabsieve}. 

Our work differs from these approaches by explicitly integrating trajectory-aware transport signals into reverse-KL optimization. Rather than only aligning local token probabilities, our framework enables token-level supervision to perceive short-horizon trajectory shifts and reasoning path dependencies, resulting in more stable and effective reasoning alignment.

\subsection{Efficient Reasoning and Agentic Systems}
Recent studies also explore efficient retrieval, reasoning, and representation learning in domain-specific settings. Existing work investigates robustness-precision trade-offs and reranking strategies in financial RAG systems~\cite{cheng2026resolvingrobustnessprecisiontradeofffinancial,cheng2026enhancingfinancialreportquestionanswering}, energy-efficient RAG architectures for small language models~\cite{cheng2026toward,xie2025chat}, and semantic embedding analysis for short-text understanding~\cite{11484350,xie2026hvd}. Related applications further include LLM-based financial disclosure analysis~\cite{liu2026improving}, co-design frameworks for efficient multimodal inference~\cite{chen2025autoneuralcodesigningvisionlanguagemodels,xie2026delving}, and time-series studies on volatility forecasting and regime-dependent market dynamics~\cite{cheng2026volatility,Cheng2026,jiang2026magma,jiang2026anatomy}, while revealing phenomena such as memory-induced behavioral instability in multi-agent environments~\cite{liu2026memory,11401016}. CoDES improves small language models by combining domain-specific LoRA fine-tuning with weighted parameter ensembling~\cite{codes2026}. Related efforts further explore structured supervision for tool-use reasoning~\cite{jiang2026scribe}, reputation-based coordination frameworks for collaborative agents~\cite{10.1007/978-981-95-4384-7_10}, and quantized multimodal systems for efficient deployment~\cite{guo2025quantized}. 

Reliable reasoning increasingly also depends on robust retrieval and evaluation mechanisms. Prior work studies LLM-as-a-judge evaluation frameworks~\cite{gao2023human,li-etal-2025-generation}, retrieval robustness under knowledge conflicts and spurious features~\cite{chen2026doesragknowretrieval,yang2025quantifying}, evidence calibration in cited RAG systems~\cite{qian2026relevantwarrantedevidenceforcecalibration}, and hybrid retrieval strategies for balancing robustness and precision~\cite{cheng2026resolvingrobustnessprecisiontradeofffinancial}. The research on LLMs' tone highlight an important reliability risk of real-world LLM deployment, supporting the need of robustness testing in high impact domain~\cite{cai2025tone}. 

Finally, reasoning systems are becoming increasingly connected with structured and graph-based information processing. Recent studies investigate graph-enhanced representations for spreadsheet understanding~\cite{lei2026sheet}, structured semantic forecasting using multiple LLM signals~\cite{zhang2026finsentllm}, scalable graph retrieval and nearest-neighbor search~\cite{wang2023towards,DBLP:journals/pvldb/WangYZLCL24}, and robustness-oriented adaptation under distribution shifts~\cite{wu2026adaptive,zeng2025enhancing}. Related domain-specific applications further demonstrate the growing use of LLM reasoning and representation learning techniques in areas such as medical image analysis, financial disclosure analysis, market behavior modeling, semantic embeddings, and interactive 3D systems~\cite{liu2026improving,dai2023analyst,dai2023neighbors,11484350,10.1117/12.3096291,li2026comprehensive}.

\section{Conclusion}

In this work, we revisit the trajectory-level assumptions underlying On-Policy Distillation for reasoning models. Although OPD is commonly expected to identify and repair reasoning errors through token-level reverse-KL supervision, our analysis shows that this assumption is only partially realized in practice. We identify two key limitations of standard OPD: (1) not all high-loss tokens correspond to real trajectory-level divergence, and many act as noisy false alarms; and (2) even when real divergent points are detected, isolated token-level correction is often insufficient to redirect the student's future reasoning trajectory.

Motivated by these observations, we propose \textbf{Trajectory Aware OPD}, which combines trajectory-aware divergence detection with short-window trajectory supervision based on optimal transport. TOPD first identifies real divergent points by measuring near-future trajectory drift, and then injects teacher future trajectory information into local training through OT-based trajectory guidance. Extensive experiments on competitive mathematics benchmarks show that TOPD consistently improves standard OPD. Further mechanistic analysis demonstrates that TOPD more effectively reduces local trajectory divergence and enables the student to better absorb teacher trajectory information during reasoning correction.

Overall, our findings suggest that reasoning failures in OPD are fundamentally trajectory-level phenomena rather than isolated token mismatches. We hope this work provides a step toward more trajectory-aware distillation objectives for reasoning language models.

\section{Limitations}

This work has several limitations. First, TOPD introduces additional computational cost because it requires short-window continuation comparison and OT computation. Second, our analysis focuses on short-horizon trajectory divergence, while some reasoning errors may emerge only over longer contexts. Third, our experiments are mainly conducted on mathematical reasoning benchmarks, so further validation is needed on other domains such as code generation and open-ended reasoning. Finally, OT distance measures trajectory proximity but does not always capture semantic equivalence, since different reasoning paths can lead to the same correct answer.

\section{Ethics}

This work uses publicly available datasets and open-access benchmark tasks. We do not access, infer, or attempt to recover any proprietary training data or internal model components. All experiments are conducted through standard inference and optimization procedures, without collecting or processing personal or sensitive user data.

\textbf{Licenses and Intended Use.}
All datasets and benchmarks are used in accordance with their released terms and intended research purposes. We do not redistribute raw datasets or proprietary model outputs. Any derived artifacts, including probing statistics and trajectory analyses, are intended only for research and evaluation.

\textbf{Artifact Documentation.}
Our experiments focus on English mathematical reasoning benchmarks such as AIME24, AIME25, and HMMT25-Feb. These artifacts primarily cover symbolic and multi-step reasoning problems rather than demographic or user-centered data.

\textbf{Risks.}
Although the datasets are publicly available and widely used, we cannot guarantee that they are entirely free from biased, toxic, or otherwise undesirable content. We use ChatGPT~\footnote{\url{https://chatgpt.com/}} by OpenAI only for grammar correction and language polishing.

\bibliography{custom}

\appendix

\section{Additional correlation analysis}
\label{correlation}
\begin{figure}[t]
    \centering
    \includegraphics[width=0.9\linewidth]{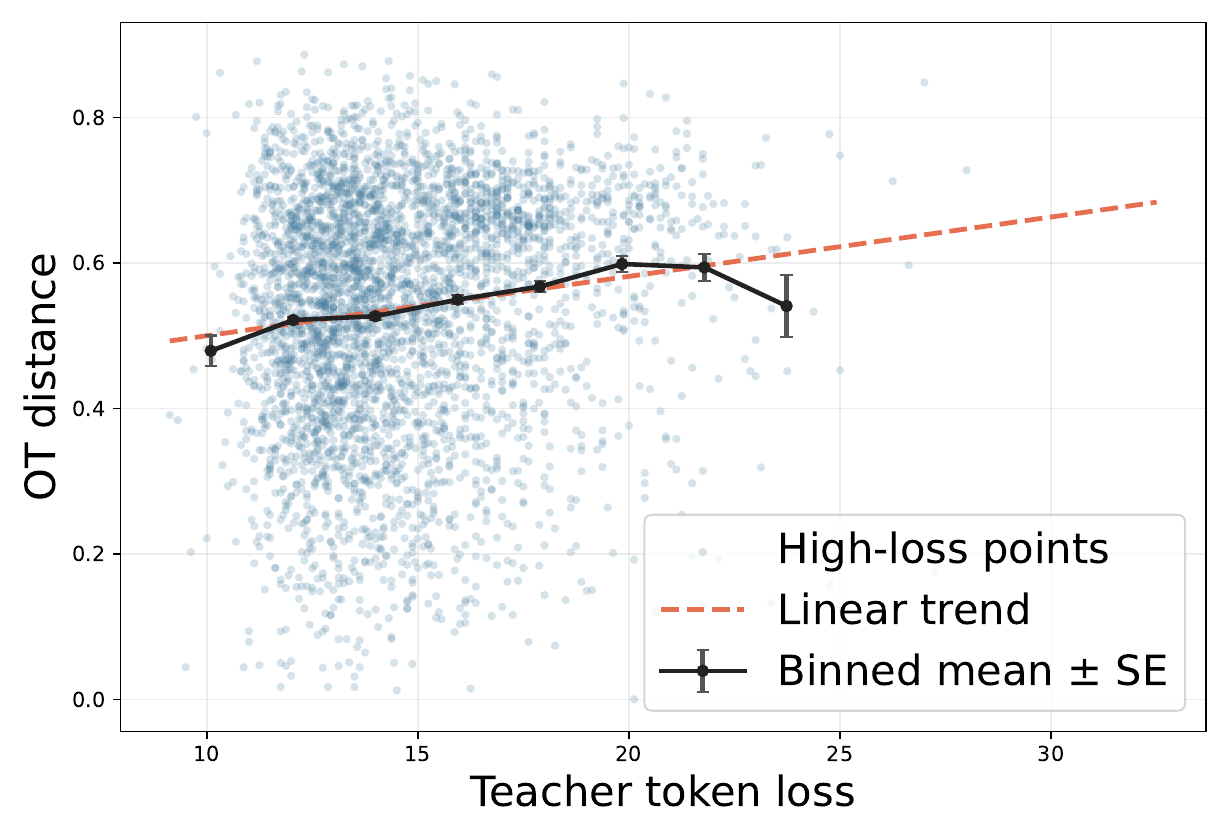}
    \caption{
    Correlation between teacher token loss and near-future OT distance for high-loss token positions. Although the overall trend is positive, the relationship remains weak and highly dispersed, indicating that token-level loss alone is an imprecise predictor of trajectory-level divergence.
    }
    \label{fig:loss_ot_scatter}
\end{figure}

To further examine the relationship between token-level loss and trajectory-level divergence, Figure~\ref{fig:loss_ot_scatter} presents the scatter distribution between teacher token loss and near-future OT distance for high-loss token positions. Although the overall linear trend is positive, the correlation remains weak, with substantial variance across the full loss range. In particular, many high-loss tokens still correspond to relatively small OT distances, while some moderate-loss tokens induce strong trajectory divergence. The binned mean trend further shows that the increase in OT distance with respect to token loss is gradual rather than sharply separable. These observations support our claim that token-level loss is informative but insufficiently selective for identifying genuine trajectory-critical divergence points.

\section{Training and Evaluation Details}
\label{app:training_details}
\subsection{off policy and on policy training}

\paragraph{Training Details.} We distill from Qwen3-30B-A3B-Instruct-2507, a Mixture-of-Experts teacher with $\sim$30B total and $\sim$3B active parameters, into Qwen3-4B-Instruct-2507 as the student, using the KDFlow framework. Both models are run with thinking mode disabled (\texttt{enable\_thinking=False}). Training proceeds in two stages on a single node of $4\times$ H100 (80\,GB) GPUs with FSDP2, bf16, and gradient checkpointing.
  
\paragraph{Stage 1 -- Off-policy KD.} We first sample 20k teacher responses on math prompts drawn from OpenThoughts3 ($\text{temperature}=0.6$, $\text{top\_p}=0.95$, $\text{max\_new\_tokens}=16384$, $\text{TP}=2$ on $2\times$ H100). The student is then trained on these (prompt, teacher-response) pairs with a per-token KL distillation loss combined with cross-entropy at $\text{kd\_ratio}=0.5$ (vanilla KD, forward KL). We use AdamW with learning rate $2\times10^{-5}$, $5\%$ linear warmup, global batch size $128$ (micro-batch $1$), max sequence length $16384$, sample packing, and ring attention of size $2$, for $1$ epoch.

\paragraph{Stage 2 -- On-policy KD.} Initialized from the Stage-1 checkpoint, the student generates $4$ rollouts per prompt ($\text{temperature}=1.0$, $\text{top\_p}=1.0$, $\text{generate\_max\_len}=8000$, $\text{prompt\_max\_len}=800$) on a separate 10k-prompt slice (positions 20k--30k of the same source), and is distilled toward the teacher's token distributions on those rollouts. We use vanilla KD with reverse KL at $\text{kd\_ratio}=1.0$ (pure distillation, no CE), learning rate $2\times10^{-6}$, $5\%$ linear warmup, gradient clipping at $1.0$, global batch size $4$ (micro-batch $1$), and $1$ epoch. The rollout engine uses $1$ engine with $\text{TP}=2$ and the teacher uses $\text{TP}=4$; both engines share GPUs with the trainer via offload-to-CPU sleep/wakeup (\texttt{teacher\_enable\_sleep=True}, \texttt{rollout\_enable\_sleep=True}.

\subsection{Computational Overhead.}
TOPD adds extra computation mainly from short-window continuation generation and OT alignment. Unlike standard OPD, which applies token-level reverse-KL supervision over the full student trajectory, TOPD only performs trajectory-level correction on selected high-loss candidate positions. Let \(N\) be the average trajectory length and let \(m\) be the number of selected high-loss positions per trajectory. The additional probing ratio is therefore approximately \(m/N\). In our setting, \(m=218\) is much smaller than \(N=9688\), so only a small fraction of positions require trajectory-level processing.

For each selected position, TOPD generates a short continuation of length \(K=50\) from the shared prefix and computes an OT alignment over the resulting \(K \times K\) cost matrix. The OT subproblem is therefore bounded by a fixed short-window size rather than the full trajectory length. As a result, the additional cost scales with \(mK\) for continuation generation and with the cost of solving \(m\) small OT problems per trajectory, rather than with all tokens in the response.

Empirically, this overhead remains manageable. Under the same training setup, TOPD increases the total training time to \(1.41\times\) that of standard OPD. This suggests that the proposed trajectory-level guidance can be incorporated into OPD with moderate additional cost.

\end{document}